\documentclass[letterpaper, 10 pt, conference]{ieeeconf}  

\IEEEoverridecommandlockouts                              

\title{\LARGE \bf
Physics-Informed Graph Learning with Uncertainty Awareness for Open-Set Domain Generalization in Fault Diagnosis 
}

\author{Jinfeng Zhu$^{1}$ and Shiyu Long$^{2}$ and Ye Yuan$^{3}$
\thanks{$^{1}$Jinfeng Zhu,  $^{2}$Shiyu Long and $^{3}$Ye Yuan are wih the College of Computer and Information Science, Southwest University, Tiansheng Road, Chongqing, 400715, China
        {\tt\small yuanyekl@swu.edu.cn}}%
}

\usepackage{amsmath}
\usepackage{amsfonts}
\usepackage[utf8]{inputenc}
\usepackage[T1]{fontenc}

\usepackage{booktabs}
\usepackage{multirow}   
\usepackage{caption}    
\usepackage{graphicx}
\graphicspath{{./images/}}

\usepackage{subcaption}

\usepackage{makecell}
\usepackage{cite}

\begin{document}

\onecolumn

\maketitle
\thispagestyle{empty}
\pagestyle{empty}

\begin{abstract}
Intelligent industrial maintenance critically relies on reliable fault diagnosis of rotating machinery. However, it faces formidable challenges from unknown fault types and domain shifts induced by varying operating conditions, which is formally formulated as the open-set domain generalization (OSDG) problem. Existing methods are mainly data-driven, thereby overlooking the cascaded propagation of uncertainty across feature extraction, topological learning, and decision-making stages. To tackle this challenge, we propose PGU-OD, a novel Physics-Informed Graph Learning framework with Uncertainty Awareness for Open-set Domain generalization. First, it designs a physics-informed spectral attention module to extract condition-robust fault features, thereby suppressing perceptual uncertainty caused by frequency shifts. Further, it constructs an uncertainty-aware adaptive graph learning mechanism to dynamically adjust the edge weights of the sample graph guided by class-scale Gaussian distribution parameters, which mitigates the structural propagation of uncertainty. Finally, a Gaussian-distribution-based adaptive boundary loss function and a dual-criteria open-set inference strategy are developed to optimize decision boundaries and reliably reject unknown faults. Extensive experimental evaluations on two public and widely used rotating machinery fault datasets demonstrate that the proposed PGU-OD outperforms state-of-the-art baselines in both known fault classification and unknown fault rejection under domain shifts. 

\end{abstract}

\section{INTRODUCTION}

The progress of Industry 4.0 has made reliable condition monitoring of rotating machinery a crucial enabler for intelligent maintenance systems \cite{liu2025adaptive,chen2023bayesian}. Deep learning has recently achieved substantial progress in the field of intelligent fault diagnosis, alongside advances in graph nerual networks \cite{wang2026advanced,G3,G4,G5,bi2025discovering} that have demonstrated remarkable capability in modeling complex relational structures and hierarchical representations \cite{wang2026graph,he2026modularized,bi2024graph,liu2023symmetry,jiang2025dual,sun2025multi,liu2024pretraining}. However, most existing methods rely on two ideal assumptions: the training and test sets share identical data distributions, and all fault categories are known during testing \cite{rehman2023open}. In practical industrial environments, these assumptions are often violated. Variations in operating conditions lead to significant domain shifts, while previously unseen fault patterns may emerge unexpectedly. Consequently, the open-set fault diagnosis (OSFD) \cite{Geng2021OSFD} task and its more difficult variant, open-set domain generalization (OSDG) , have become critical research topics in intelligent maintenance \cite{Xu2026fosfd,li2025openbogie}, \cite{chen2023transfer}.

Recently, representative approaches for open-set and cross-domain diagnosis include boundary-based methods (OpenMax \cite{bendale2016towards}, EVT \cite{yu2021deep}), adversarial domain adaptation (EFECANN  \cite{Geng2025efecann}, dual-adversarial networks \cite{zhao2022dual,lei2025fault,bi2025open}), generative models with auxiliary classifiers \cite{liao2025novel,jin2024few}, and advanced OSDG framewokrs (MDCC \cite{lu2024novel}, ACDPN \cite{jia2025auxiliary,fu2025reason,fu2023magva,xiao2025uncertainty,an2025certainty,xu2025novel,chen2023open}). PID-based optimization \cite{G6}, \cite{G7} and Kalman-filter-based uncertainty estimation \cite{G8} also enhance robustness under dynamic conditions. , and recent extensions have further integrated adaptive divergence strategies and fuzzy logic control to improve convergence and adaptability in complex industrial scenarios \cite{yuan2025proportional,yuan2020temporal,yuan2024fuzzy,yuan2023adaptive}.

Despite these advances, most existing approaches remain purely data-driven and overlook the cascaded propagation of uncertainty across different stages of the diagnosis process. First, the feature extraction stage typically lacks physical constraints, making learned representations sensitive to frequency shifts caused by varying operating conditions \cite{li2022explainable}, which introduces perceptual uncertainty, and undermines the reliability of subsequent topological learning and decision-making processes \cite{qin2026robust,hu2025comprehensive,chen2024latent,wu2023robust,chen2024generalized,he2023physics,li2025deep}. Second, graph-based models often ignore the heterogeneity of sample confidence during topology construction, allowing highly uncertain samples near decision boundaries to contaminate the entire graph structure \cite{wang2024distributed,wu2026multimetric,lyu2025dynamic,wan2024self,jia2024causal}. Finally, at the decision stage, existing methods lack mechanisms that tightly couple physical features with statistical distributions, making it difficult to adapt to domain-induced variations in intra-class compactness and resulting in unreliable open-set rejection boundaries~\cite{lundgren2022data}.

To address these critical challenges, this paper uniquely proposes PGU-OD, a Physics-Informed Graph Learning Framework with Uncertainty Awareness. PGU-OD establishes an end-to-end probabilistic diagnostic closed-loop to systematically mitigate multi-layer uncertainty from feature extraction to final decision-making, leveraging insights from adaptive latent factor analysis and nonnegative tensor factorization for robust high-dimensional data representation \cite{yuan2026novel,shang2018randomized,lin2025ncsac,wu2023mmlf,xu2025attention,he2025individual,qian2025adaptive}. The key contributions of this work are threefold: 

\begin{enumerate}
    \item A physics-informed feature extraction framework (PISA-Net) is proposed, which integrates frequency-constrained wavelet convolution and spectral attention to address feature degradation caused by fault frequency drift under varying operating conditions.
    \item An uncertainty-aware adaptive graph network is constructed. By utilizing class-scale parameters to guide edge weighting, it forms a closed-loop coupling between structural learning and uncertainty estimation to effectively mitigate ambiguous information propagation. 
    \item A Gaussian-distribution-based adaptive boundary joint loss and dual open-set decision criterion are designed to enable end-to-end uncertainty management, achieving superior performance in cross-domain open-set fault diagnosis scenarios.
\end{enumerate}

\section{Methodology}

\subsection{Problem Formulation}

\begin{figure}[htbp]
   \centering
   \includegraphics[width=0.8\columnwidth]{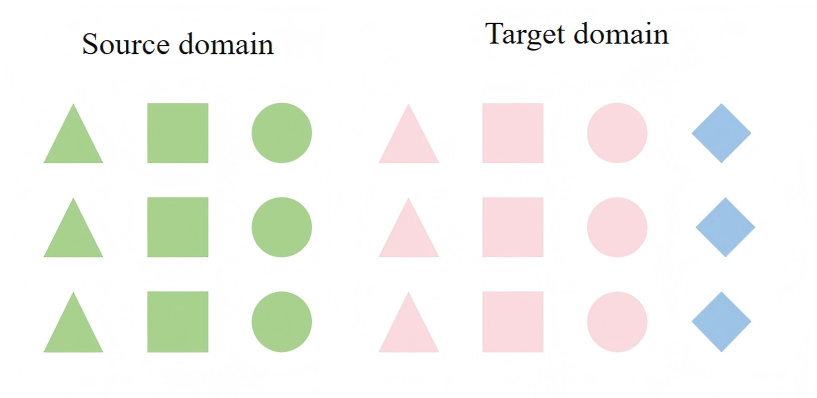}
   \caption{classes between source and target.}
   \label{fig:S_T_C}
\end{figure}

Let the labeled source-domain dataset be denoted as $\mathcal{D}_s = \{ (\mathbf{x}_i^s, y_i^s) \}_{i=1}^{N_s}$, $y_i^s \in \mathcal{Y}_k$ and the unlabeled target-domain dataset as $\mathcal{D}_{t}=\left\{\mathbf{x}_{j}^{t}\right\}_{j=1}^{N_{t}}$. Here, $\mathcal{Y}_k \subset \mathcal{Y}_t$ represents the set of known classes, while the unknown-class set is defined as $\mathcal{Y}_u = \mathcal{Y}_t \setminus \mathcal{Y}_k$, as illustrated in  Fig.~\ref{fig:S_T_C}. Furthmore, due to variations in operating conditions such as speed and load, the source and target domains follow different data distributions, leading to domain shift $P_s(\mathbf{x}) \neq P_t(\mathbf{x})$, and the objective of cross-domain open-set fault diagnosis is to learn a feature mapping:
\begin{equation}
    f_\theta : \mathcal{X} \to \mathbb{R}^d
    \label{eq:feature mapping}
\end{equation}
that minimizes classification errors for known classes while enabling reliable rejection of samples belonging to unknown classes, which aligns with the broader objective of learning compact and well-separated feature spaces in high-dimensional and incomplete data scenarios \cite{li2025neural,qin2023asynchronous}.

\begin{figure}[htbp] 
    \centering
    \includegraphics[width=0.95\textwidth]{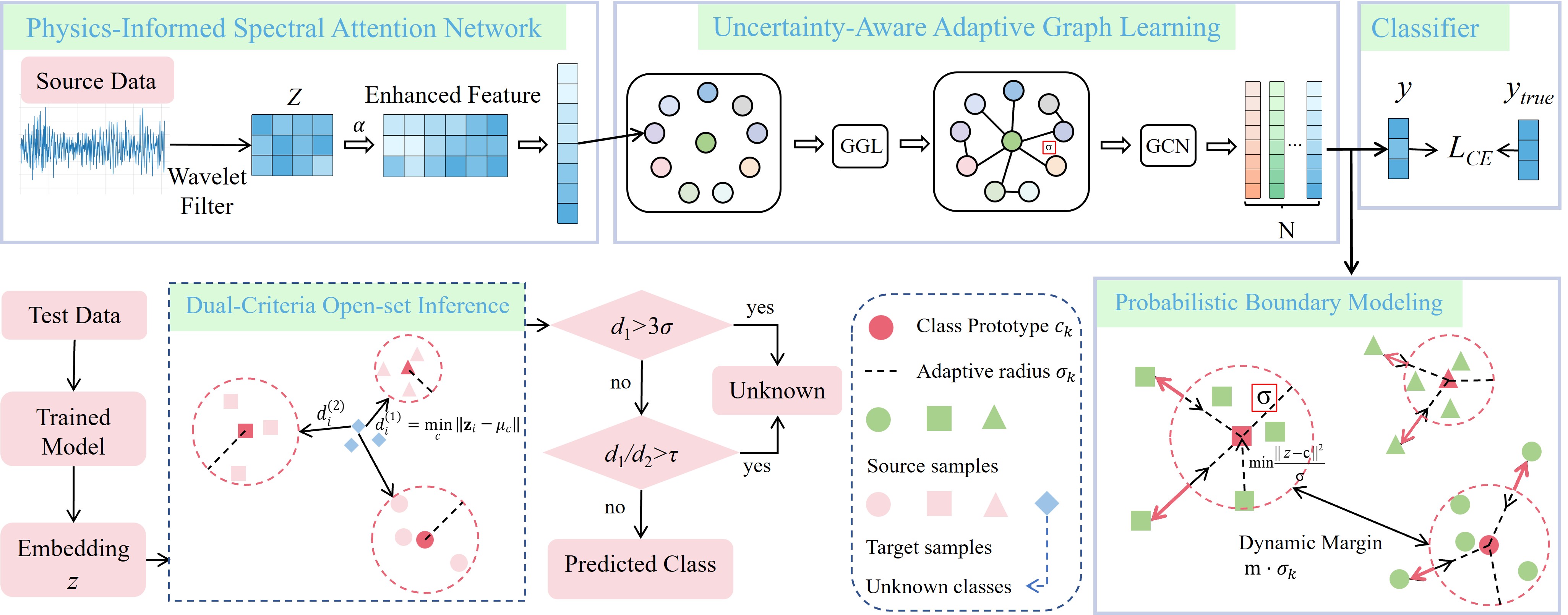} %
    \caption{The overall framework of the proposed PGU-OD.}
    \label{fig:framework}
\end{figure}

To tackle the aforementioned OSFD challenges, we propose the PGU-OD framework. As illustrated in Fig.~\ref{fig:framework}, the overall architecture consists of three tightly coupled modules designed to mitigate the cascaded propagation of uncertainty. First, raw vibration signals are fed into the PISA-Net to extract physics-informed and condition-robust feature representations, fundamentally suppressing perceptual uncertainty. Second, these representations are utilized to construct an Uncertainty-Aware Adaptive Graph, where class-level scale parameters adaptively adjust the heteroscedastic kernel bandwidths to prevent the topological propagation of ambiguous information. Finally, the updated node embeddings are optimized via a Gaussian Adaptive Boundary module, which jointly learns discriminative class prototypes and adaptive decision radii, providing a statistical basis for the dual-criteria open-set inference.

\begin{figure}[htbp]  
   \centering
   \includegraphics[width=\linewidth]{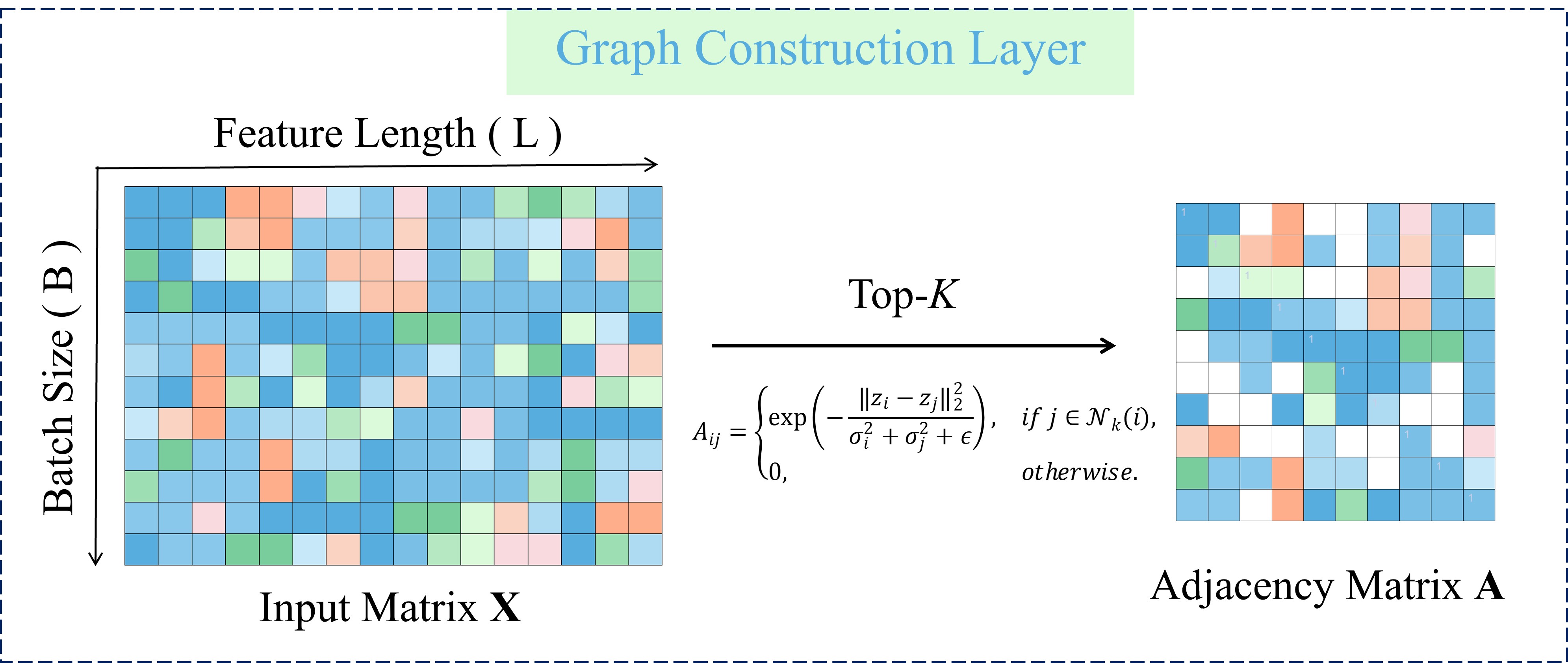}
   \caption{Illustration of the Graph Construction Layer (GCL).}
   \label{fig:GCL}
\end{figure}

\subsection{Physics-Informed Spectral Attention Network (PISA-Net) }

To extract physically interpretable and condition-robust representations, we design a Physics-Informed Spectral Attention Network (PISA-Net). The key idea is to replace the first convolutional layer of conventional CNNs with a bank of learnable wavelet filters under physical constraints, inspired by \cite{G9,G10,shao2025lsfconvformer,tang2025complex}, and then introduce a spectrum-aware attention mechanism to adaptively focus on informative frequency bands. 

\subsubsection{Constrained Learnable Wavelet Convolution}

Conventional CNN filters lack explicit physical meaning and are therefore sensitive to frequency shifts caused by variations in rotational speed \cite{lyu2026gnetic,jiang2024iterative,qin2024parallel,he2023physics,magadan2024explainable}. To address this issue, the filters in the first convolutional layer are parameterized using the Morlet wavelet form.

For the $k$-th filter, the discrete wavelet kernel is defined as
\begin{equation}
    w_k(t; s_k, \xi_k) = C_k \exp\left(-\frac{t^2}{2 s_k^2}\right) \cos\left(2 \pi \xi_k t\right)
    \label{eq:wavelet kernel}
\end{equation}
where $s_{k}>0$ denotes the scale parameter, $\xi_{k} \in[0,0.5]$ is the center frequency constrained by the Nyquist limit, and $C_{k}$ is a normalization constant.

Unlike traditional wavelet transforms with fixed parameters, the filter bank in PISA-Net is learnable, allowing the network to adaptively adjust the spectral distribution through backpropagation to match the frequency characteristics of fault signatures. . This adaptive mechanism shares conceptual similarities with latent factor models that dynamically update their parameters to capture evolving data patterns \cite{bi2024fast,liao2025local,xu2026sampling,lyu2025dynamic}.

To ensure differentiability and maintain valid parameter ranges, the parameters are re-parameterized as $\xi_k=0.5 \cdot S(\theta_k^f)$, $s_k=\text{Softplus}(\theta_k^s)+s_0$, where $S(\cdot)$ denotes the Sigmoid activation function, $\theta_k^f$ and $\theta_k^s$ are learnable parameters, $s_0>0$ prevents degenerate solutions.

Given an input signal $\mathbf{x} \in \mathbb{R}^{B \times 1 \times L}$, the convolution output of the $k$-th filter is $z_k=x \ast w_k$. To enhance impulsive fault signatures, the absolute value is taken to approximate the envelope response. The outputs of all $K$ filters form the spectral feature map $\mathbf{Z} \in \mathbb{R}^{B \times K \times L}$.

\subsubsection{Spectrum-Aware Attention Mechanism}

After obtaining the multi-band feature map $Z$, we introduce a spectrum-aware attention mechanism to adaptively emphasize informative frequency bands and suppress irrelevant noise components.

First, we combine GAP and GMP to capture steady-state energy and transient impulse characteristics~\cite{tang2025nerual,wei2022robust,chaouech2025advanced}. The band-wise descriptor is computed as\begin{equation}
    u_k = \mathrm{GAP}(Z_k) + \mathrm{GMP}(Z_k)
    \label{eq:statistical description}
\end{equation}

Next, a lightweight one-dimensional convolution layer is used to learn inter-band dependencies and generate attention weights $\alpha=\sigma(\text{Conv1D}(u))$, where $\sigma(\cdot)$ denotes the Sigmoid activation function that constrains the weights within the range $(0,1)$.

Finally, the attention weights are applied, $\tilde{Z}_{k,l}=\alpha_k Z_{k,l}$ to recalibrate the spectral feature map. The recalibrated feature representation $\mathbf{\tilde{Z}}$ enhances frequency components that are physically relevant to fault signatures, providing more robust features for subsequent processing.

\subsection{Uncertainty-Aware Adaptive Graph Learning}

To model relationships among samples while suppressing structural uncertainty caused by ambiguous samples, we propose an adaptive graph learning strategy guided by class scale parameters~\cite{jiang2025dual,chang2025causal}. This approach is motivated by recent studies on graph neural networks that emphasize the importance of adaptive topology learning and uncertainty-aware aggregation in complex data environments \cite{wu2026non,tang2025auto,wang2025convolution,wu2024fine}.

\subsubsection{Gaussian Class Modeling} 

We assume that feature embeddings follow an isotropic Gaussian distribution conditioned on class label $c$:
\begin{equation}
p(\mathbf{z} \mid y=c) = \mathcal{N}(\boldsymbol{\mu}_c, \sigma_c^2 \mathbf{I})
\label{eq:gaussian_model}
\end{equation}
where $\boldsymbol{\mu}_c$ represents the class prototype, and $\sigma_c$ is a learnable scale parameter reflecting the intra-class dispersion.

The scale parameter is defined as
\begin{equation}
\sigma_c = \text{Softplus}(\rho_c) + \epsilon_0
\label{eq:scale_parameter}
\end{equation}
where $\rho_c$ is an unconstrained learnable parameter, and $\epsilon_0$ is a small positive constant for numerical stability.

\subsubsection{Uncertainty Feedback via Scale Parameters}

For an arbitrary sample embedding $\mathbf{z}_i$, its predicted class label is determined by the nearest class prototype:
\begin{equation}
y_i = \underset{c}{\arg\min} \left\| \mathbf{z}_i - \boldsymbol{\mu}_c \right\|_2^2
\label{eq:predicted_label}
\end{equation}
We define the uncertainty estimate for this sample as the scale parameter of its predicted class $\sigma_i = \sigma_{y_i}$, which design propagates global class statistics to the individual sample level, enabling uncertainty-aware graph construction.

\subsubsection{$\sigma$-Weighted Adjacency Matrix}

In the feature space $\mathbf{Z}$, for each node $i$, we establish a $K$-nearest neighbor (KNN) graph, denoted as $\mathcal{N}_k(i)$.

Instead of assigning equal weights to edges, the proposed method incorporates sample uncertainty to adjust edge strengths. The adjacency weight between node $i$ and node $j$ is defined as
\begin{equation}
A_{ij} =
\begin{cases}
\exp\left(-\frac{\| \mathbf{z}_i - \mathbf{z}_j \|^2}{\sigma_i^2 + \sigma_j^2 + \epsilon}\right), & j \in \mathcal{N}_k(i) \\
0, & \text{otherwise}
\end{cases}
\label{eq:adjacency_weight}
\end{equation}

This heteroscedastic Gaussian kernel adapts its bandwidth via the uncertainty ($\sigma$) of each sample pair. A node exhibits high uncertainty (i.e., a larger $\sigma_i$), its similarity kernel broadens, producing uniformly higher pre-normalized edge weights ($A_{ij} \to 1$) within its $K$-neighborhood \cite{yang2025link,deng2026fuzzy,sun2025multi,yan2025compact}. Crucially, during the subsequent symmetric normalization ($\mathbf{A} = \mathbf{D}^{-1/2} \mathbf{A} \mathbf{D}^{-1/2}$), this dense connectivity increases the uncertain node's degree $D_{ii}$, diluting its message-passing weights by $1/\sqrt{D_{ii}}$). Rather than severing connections, this diffusion mechanism prevents highly uncertain samples from exerting concentrated \cite{qin2024ada}, dominant influence over specific neighbors, thereby stabilizing the global topology and mitigating ambiguous structural propagation. 

Finally, a graph convolution layer is applied to aggregate neighborhood information and update node embeddings $\mathbf{Z}' = \tilde{\mathbf{A}} \mathbf{Z} \mathbf{W}$, where $\mathbf{W} \in \mathbb{R}^{d \times d}$ is a learnable weight matrix, , seamlessly integrating structural contexts while remaining robust to uncertain nodes. 

\subsection{Gaussian Adaptive Boundary Optimization}

We adopt supervised contrastive loss $L_{\text{SupCon}}$ \cite{khosla2020supervised,wan2024self,jia2024causal} to enhance feature discrimination.

\subsubsection{Boundary-Aware Log-Likelihood} 

Based on the Gaussian distribution assumption, we design a boundary-aware loss consisting of intra-class and inter-class terms.

The intra-class term minimizes the Mahalanobis distance between each sample and its corresponding class prototype while preventing the scale parameter from collapsing~\cite{yan2025compact,shao2025safs}:
\begin{equation}
L_{\text{intra}} = \frac{1}{N} \sum_i \left( \frac{\| \mathbf{z}_i - \boldsymbol{\mu}_{y_i} \|^2}{\sigma_{y_i}^2} + \log \sigma_{y_i} \right)
\label{eq:loss_intra}
\end{equation}

The inter-class term enforces a dynamic margin against the nearest incorrect prototype~\cite{shao2021novel,shao2021intelligent}:
\begin{equation}
L_{\text{inter}} = \frac{1}{N} \sum_i \max\left(0, d_i^{(1)} + m \sigma_{y_i} - d_i^{(2)}\right)
\label{eq:loss_inter}
\end{equation}
where $d_i^{(1)}$ and $d_i^{(2)}$ are distances to the nearest and second-nearest prototypes. This adaptive margin allows classes with larger dispersion  ($\sigma_c$)  to maintain proportionally larger decision regions. 

\subsubsection{Overall Training Objective}

The total training objective is defined as the weighted combination of the above losses:
\begin{equation}
L_{\text{total}} = L_{\text{CE}} + \lambda_1 L_{\text{SupCon}} + \lambda_2 (L_{\text{intra}} + L_{\text{inter}})
\label{eq:loss_total}
\end{equation}
where $L_{\text{CE}}$ is the standard cross-entropy loss ensuring classification accuracy, and $\lambda_1$ and $\lambda_2$ balance the contributions of different objectives. Through joint optimization, the model learns an embedding space that is both discriminative and statistically interpretable.

\subsection{Dual-Criteria Open-Set Inference}

During testing, two distances are defined: $d_{i}^{(1)}=\min _{c}\left\|\mathbf{z}_{i}-\boldsymbol{\mu}_{c}\right\|$, and $d_{i}^{(2)}$ is the second smallest distance~\cite{zhao2020deep,li2025openset}.

Two complementary criteria are used for open-set detection: The relative criterion, $\frac{d_i^{(1)}}{d_i^{(2)}} > \tau_r \Rightarrow \text{Unknown}$, and the absolute statistical criterion, $d_i^{(1)} > \kappa \sigma_{y_i} \Rightarrow \text{Unknown}$.

By combining relative distance comparison and absolute statistical thresholds, the proposed approach achieves reliable rejection of unknown faults under cross-domain conditions.

\section{EXPERIMENTS AND DISCUSSION}

\subsection{Experiment Setup}

In the OSFD setting, the source domain exclusively contains known classes $Y_k$, while the target domain comprises $Y_k \cup Y_u$.  All vibration signals are segmented into samples of length 1024 using a sliding window approach. We conducted experiments on two publicly available rotating machinery datasets:

\subsubsection{Case Western Reserve University Bearing Dataset}
CWRU \cite{smith2015rolling} contains seeded faults of varying sizes (inner race, outer race, and ball faults) created via electro-discharge machining. With three fault diameters considered—0.007, 0.014, and 0.021 inches—the dataset includes 10 operating states: one normal and nine faulty conditions. The sampling frequency is 12 kHz. The specific open-set task settings are detailed in Table \ref{tab:cwru_dataset} and Table \ref{tab:cwru_tasks}.

\begin{table}[htbp]
\centering
\caption{CWRU Bearing Dataset Fault Types and Sizes}
\label{tab:cwru_dataset}
\resizebox{\columnwidth}{!}{
\begin{tabular}{@{}lccccccccccc@{}} 
\toprule
\multirow{2}{*}{\textbf{Dataset}} & \multirow{2}{*}{\textbf{Class}} & \multirow{2}{*}{\textbf{NC}} & \multicolumn{3}{c}{\textbf{Inner Race (IR)}} & \multicolumn{3}{c}{\textbf{Ball (B)}} & \multicolumn{3}{c}{\textbf{Outer Race (OR)}} \\ \cmidrule(lr){4-6} \cmidrule(lr){7-9} \cmidrule(l){10-12}
 &  &  & IR1 & IR2 & IR3 & B1 & B2 & B3 & OR1 & OR2 & OR3 \\ \midrule
\multirow{2}{*}{CWRU} & Fault Size (mil) & 0 & 7 & 14 & 21 & 7 & 14 & 21 & 7 & 14 & 21 \\ \cmidrule(l){2-12} 
 & Fault Type Label & 0 & 1 & 2 & 3 & 4 & 5 & 6 & 7 & 8 & 9 \\ \bottomrule
\end{tabular}
}
\end{table}

\begin{table}[htbp]
\centering
\caption{Experiment Setup of CWRU Dataset}
\label{tab:cwru_tasks}
\small 
\setlength{\tabcolsep}{10pt} 
\begin{tabular}{@{}ccccc@{}}
\toprule
\textbf{Task} & \textbf{Source (rpm)} & \textbf{Target (rpm)} & \textbf{Source Class} & \textbf{Target Class} \\
\midrule
P1& 1797 & 1772 & 0,1,2,3 & 0,1,2,3,4,5,6 \\
P2& 1797 & 1750 & 0,1,4,5,6 & 0,1,3,4,5,6,7 \\
P3& 1772 & 1750 & 0,1,3,4,7 & 0,1,2,3,4,5,6,7,8 \\
P4& 1772 & 1730 & 0,4,5,6 & 0,1,2,3,4,5,6,7,8,9 \\
P5& 1750 & 1772 & 0,1,2,3,4,5,6 & 0,1,2,3,4,5,6,7,8,9 \\
P6& 1750 & 1730 & 0,1,2,3,4,7,8,9 & 0,1,2,3,4,5,6,7,8,9 \\
P7& 1730 & 1797 & 0,7,8 & 0,3,4,7,8,9 \\
P8& 1730 & 1772 & 0,1,3,4,6,7 & 0,1,2,3,4,5,6,7,8 \\
\bottomrule
\end{tabular}
\end{table}

\subsubsection{Paderborn Dataset}The PU dataset \cite{lessmeier2016condition} employs 6203-type deep groove ball bearings and adopts a sampling rate of 64 kHz. The fault categories cover normal condition, inner race fault, outer race fault, and combined fault. The experimental setup are listed as shown in Table \ref{tab:paderborn_tasks}.

\begin{table}[htbp]
\centering
\caption{Experiment Setup of Paderborn Dataset}
\label{tab:paderborn_tasks}
\begin{tabular}{@{}c c c@{}}
\toprule
\multicolumn{3}{c}{Paderborn Dataset} \\
\midrule
Task & Source Conditions & Target Conditions \\
\midrule
P1 & 1500rpm -- 0.7Nm -- 400N & 1500rpm -- 0.1Nm -- 1000N \\
P2 & 1500rpm -- 0.1Nm -- 1000N & 1500rpm -- 0.7Nm -- 400N \\
P3 & 1500rpm -- 0.7Nm -- 400N & 1500rpm -- 0.7Nm -- 1000N \\
P4 & 1500rpm -- 0.7Nm -- 1000N & 1500rpm -- 0.7Nm -- 400N \\
P5 & 1500rpm -- 0.1Nm -- 1000N & 1500rpm -- 0.7Nm -- 1000N \\
P6 & 1500rpm -- 0.7Nm -- 1000N & 1500rpm -- 0.1Nm -- 1000N \\
P7 & 900rpm -- 0.7Nm -- 1000N & 1500rpm -- 0.7Nm -- 400N \\
P8 & 900rpm -- 0.7Nm -- 1000N & 1500rpm -- 0.1Nm -- 1000N \\
\bottomrule
\end{tabular}
\par\vspace{2pt}
\footnotesize \textit{Note:} For all tasks, the source classes are \{Normal, Inner, Outer\}, and the target classes additionally include \{Composite\} faults as unknowns.
\end{table}

In the PGU-OD, the number of wavelet filters is set to $K=32$, and their center frequencies are initialized uniformly. The number of neighbors for graph construction is set to $k=10$. The total loss function is given by \eqref{eq:loss_total}, where $\lambda_1 = 0.5$ and $\lambda_2 = 0.1$. For the dual-criteria open-set inference, the relative distance threshold is empirically set to $\tau_r = 0.3$ based on the parameter sensitivity analysis, and $\kappa=3$ follows the Gaussian $3\sigma$ principle.

The Adam optimizer is employed with an initial learning rate of $1 \text{e-}3$, and the model is trained for 100 epochs.

\subsection{Comparison Methods}

To evaluate the superiority of PGU-OD, we benchmark it against six representative methods. These include a classical OSDA method (OSBP \cite{saito2018open}) using adversarial backpropagation; DW-DAAN \cite{jian2024open}, which combines ACGAN-generated pseudo-unknowns with dual-level weighting; and KTR-BUNN \cite{wang2025knowledge}, a bi-unbiased network featuring an uncertainty-aware transferability evaluator and semantic alignment. Furthermore, we compare against three advanced OSDG frameworks: AOSDGN \cite{zhao2022adaptive}, employing triplet-loss-based local clustering and adaptive boundaries; MDCC \cite{lu2024novel}, which utilizes multi-domain contrastive coding for domain-invariant representations; and ACDPN \cite{jia2025auxiliary}, which integrates causal intervention and a memory dynamic penalty to improve sensitivity to unknown classes. 

\subsection{Evaluation Metrics}

The H-score is used to comprehensively evaluate both known class identification and unknown class rejection capabilities, serving as the core metric for this task.
    \begin{equation}
    H = \frac{2 \cdot Acc_{\text{known}} \cdot Acc_{\text{unknown}}}{Acc_{\text{known}} + Acc_{\text{unknown}}}
    \label{eq:h_score}
    \end{equation}

Its harmonic form penalizes imbalance between known accuracy and unknown rejection, reflecting practical safety and cost trade-offs. . 
    
\subsection{Overall Performance under OSFD}

\begin{table}[t]
\centering
\caption{Experimental Results on the CWRU Dataset}
\label{tab:cwru_results}
\resizebox{\columnwidth}{!}{
\begin{tabular}{c c c c c c c c}
\toprule
Task & OSBP & DW-DANN & KTR-BUNN & AOSDGN & MDCC & ACDPN & \textbf{Ours} \\
\midrule
P1 & 83.25$\pm$0.82 & 93.85$\pm$0.45 & 94.40$\pm$0.51 & 88.62$\pm$0.74 & 81.72$\pm$0.96 & 94.14$\pm$0.48 & \textbf{94.77}$\pm$\textbf{0.47}\\
P2 & 87.86$\pm$0.75 & 96.78$\pm$0.33 & 98.23$\pm$0.28 & 80.22$\pm$1.05 & 89.89$\pm$0.62 & 96.39$\pm$0.41 & \textbf{98.50}$\pm$\textbf{0.18} \\
P3 & 77.34$\pm$1.25 & 91.96$\pm$0.65 & 93.63$\pm$0.42 & 89.14$\pm$0.88 & 83.11$\pm$1.12 & 91.34$\pm$0.57 & \textbf{95.71}$\pm$\textbf{0.25} \\
P4 & 74.79$\pm$1.34 & 94.53$\pm$0.49 & 95.58$\pm$0.38 & 90.54$\pm$0.76 & 77.86$\pm$1.20 & 94.08$\pm$0.42 & \textbf{96.07}$\pm$\textbf{0.15} \\
P5 & 75.83$\pm$1.18 & 94.56$\pm$0.52 & 95.54$\pm$0.44 & 81.54$\pm$1.02 & 84.01$\pm$0.95 & 94.73$\pm$0.39 & \textbf{95.58}$\pm$\textbf{0.22} \\
P6 & 75.72$\pm$1.22 & 94.38$\pm$0.50 & 95.04$\pm$0.46 & 80.92$\pm$1.15 & 84.99$\pm$0.88 & 93.89$\pm$0.55 & \textbf{95.17}$\pm$\textbf{0.59}\\
P7 & 74.19$\pm$1.45 & 93.05$\pm$0.62 & 93.33$\pm$0.58 & 83.16$\pm$0.98 & 82.11$\pm$1.05 & 92.28$\pm$0.45 & \textbf{94.55}$\pm$\textbf{0.28} \\
P8 & 77.31$\pm$1.10 & 93.59$\pm$0.55 & 94.94$\pm$0.40 & 85.63$\pm$0.82 & 83.42$\pm$0.90 & 93.57$\pm$0.61 & \textbf{95.23}$\pm$\textbf{0.20} \\
\bottomrule
\end{tabular}
}
\end{table}

\begin{table}[t]
\centering
\caption{Experimental Results on the PU Dataset}
\label{tab:pu_results}
\resizebox{\columnwidth}{!}{
\begin{tabular}{c c c c c c c c}
\toprule
Task & OSBP & DW-DANN & KTR-BUNN & AOSDGN & MDCC & ACDPN & \textbf{Ours} \\
\midrule
P1 & 80.40$\pm$0.95 & 94.25$\pm$0.48 & 96.86$\pm$0.32 & 86.46$\pm$0.77 & 81.13$\pm$1.02 & 97.55$\pm$0.28 & \textbf{97.86}$\pm$\textbf{0.15} \\
P2 & 80.80$\pm$0.92 & 86.92$\pm$0.74 & 84.86$\pm$0.85 & 82.29$\pm$0.98 & 80.92$\pm$1.10 & 88.47$\pm$0.66 & \textbf{92.16}$\pm$\textbf{0.26} \\
P3 & 62.92$\pm$1.55 & 61.06$\pm$1.62 & 65.07$\pm$1.48 & 53.31$\pm$1.85 & 60.93$\pm$1.68 & 64.37$\pm$1.50 & \textbf{79.68}$\pm$\textbf{1.55}\\
P4 & 71.77$\pm$1.35 & 84.53$\pm$0.88 & 83.08$\pm$0.92 & 83.18$\pm$0.90 & 76.04$\pm$1.25 & 87.27$\pm$0.72 & \textbf{88.22}$\pm$\textbf{0.42} \\
P5 & 80.41$\pm$0.96 & 84.56$\pm$0.84 & 84.68$\pm$0.80 & 81.54$\pm$1.05 & 75.67$\pm$1.30 & 88.06$\pm$0.65 & \textbf{96.89}$\pm$\textbf{0.22} \\
P6 & 82.44$\pm$0.85 & 86.38$\pm$0.76 & 93.77$\pm$0.45 & 79.29$\pm$1.12 & 78.33$\pm$1.18 & 97.59$\pm$0.33 & \textbf{98.83}$\pm$\textbf{0.25}\\
P7 & 73.21$\pm$1.42 & 79.05$\pm$1.10 & 83.61$\pm$0.95 & 72.14$\pm$1.50 & 72.79$\pm$1.48 & 85.77$\pm$0.82 & \textbf{87.35}$\pm$\textbf{0.48} \\
P8 & 83.64$\pm$0.88 & 95.59$\pm$0.40 & 94.22$\pm$0.52 & 82.49$\pm$0.94 & 75.45$\pm$1.22 & 98.26$\pm$0.25 & \textbf{98.75}$\pm$\textbf{0.12} \\
\bottomrule
\end{tabular}
}
\end{table}

Table \ref{tab:cwru_results} and \ref{tab:pu_results} present the performance comparison of all methods under the OSFD setting. The proposed PGU-OD achieves the best results on most transfer tasks, demonstrating its effectiveness in cross-domain open-set fault diagnosis. OSBP, which follows the OSDA paradigm and leverages target-domain data during training, performs competitively on some tasks but shows notable performance fluctuations. Recent OSDG methods learn domain-invariant representations without accessing target data, providing a stronger comparison baseline. While these methods yield relatively competitive results, their performance degrades significantly on challenging tasks, where domain discrepancy and fault distribution shifts are more severe.In contrast, PGU-OD consistently outperforms or matches all baselines across both datasets, with particularly notable gains on the difficult tasks. This improvement stems from its synergistic design: physics-informed spectral extraction, uncertainty-aware adaptive graph modeling, and Gaussian adaptive boundary learning collectively provide robust representations and reliable decision boundaries under severe domain shifts.

\subsection{Ablation Study} 

\begin{figure*}[!htbp] 
   \centering
   \begin{subfigure}[b]{0.48\textwidth}
       \centering
       \includegraphics[width=\linewidth]{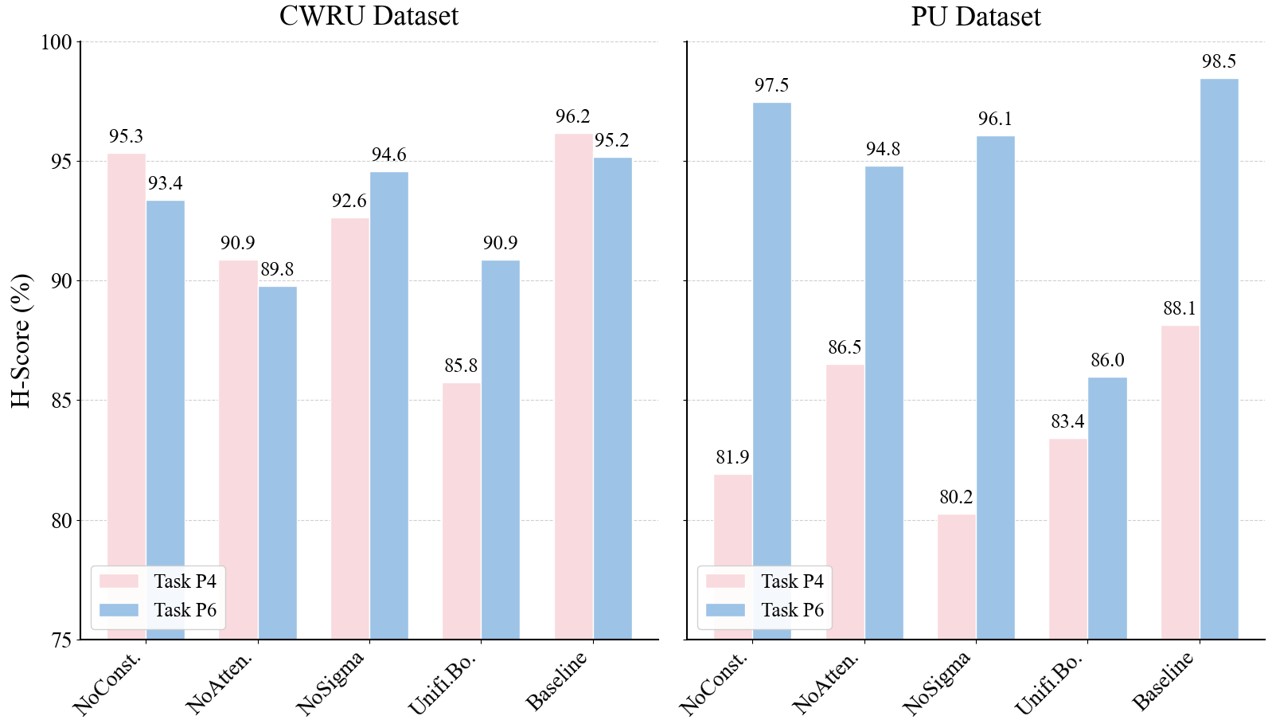}
       \caption{Ablation study results}
       \label{fig:ablation}
   \end{subfigure}
   \hfill 
   \begin{subfigure}[b]{0.48\textwidth}
       \centering
       \includegraphics[width=\linewidth]{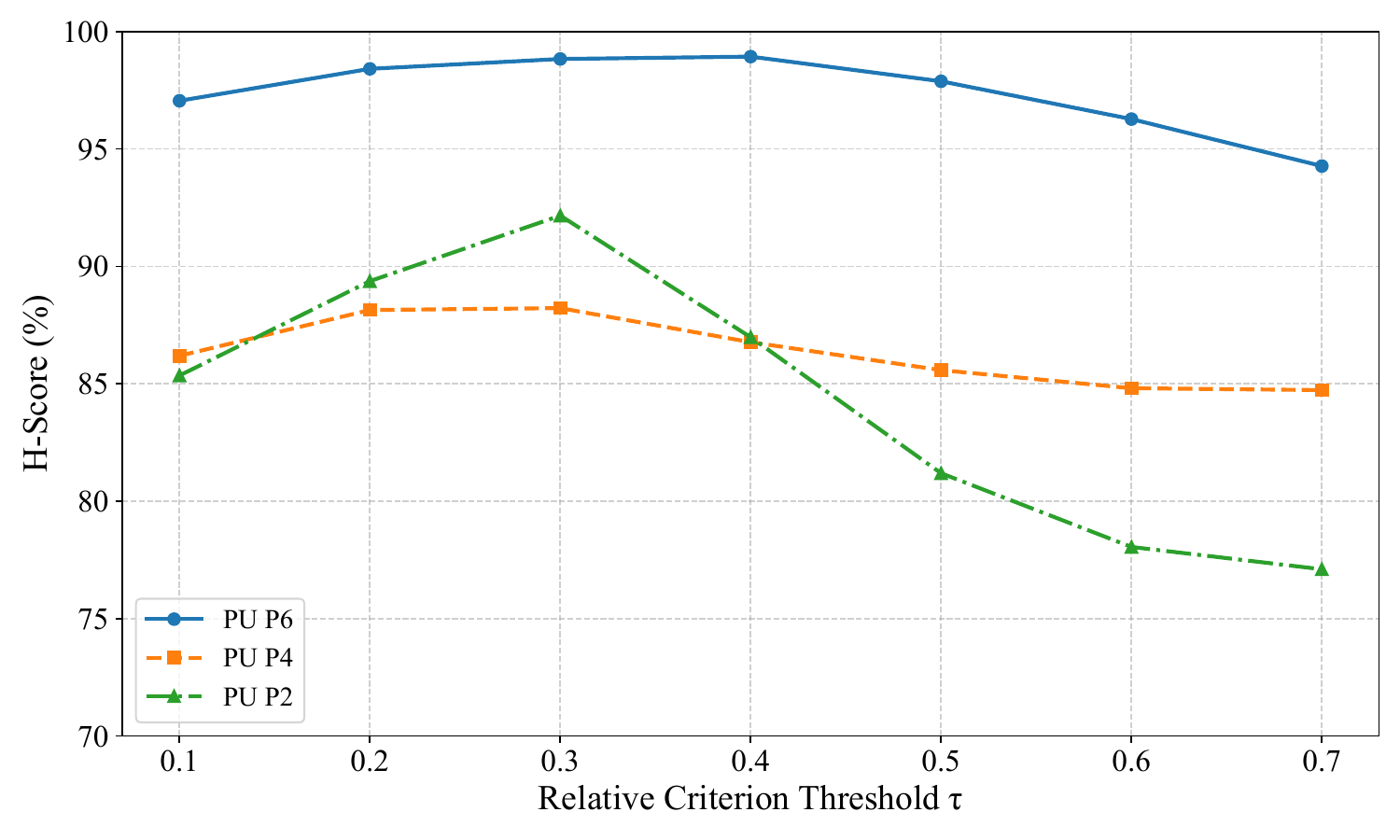}
       \caption{Impact of threshold ($\tau_r$)}
       \label{fig:relative criterion}
   \end{subfigure}
   
   \caption{Experimental evaluation and parameter sensitivity analysis.}
   \label{fig:overall_experimental_results}
\end{figure*}

To investigate the individual contributions of the proposed modules, we designed four ablation configurations. The results across four transfer tasks (P4 and P6 on both CWRU and PU datasets) are visualized in Fig. \ref{fig:ablation}, where Baseline denotes the complete PGU-OD model.

\subsubsection{Physics-Informed Features}
Replacing the wavelet convolution with a standard Conv1D layer (denoted as 'NoConst.') reduces the H-Score, validating that frequency-domain priors enhance feature stability. Further removing the spectrum-aware attention (denoted as 'NoAtten.') causes a more significant decline, underscoring its necessity in adaptively highlighting informative frequency bands. 

\subsubsection{Uncertainty-Aware Graph}
Disabling the uncertainty-guided edge weighting (denoted as 'NoSigma.') degrades the model to a standard KNN graph. The consequent performance drop proves that incorporating sample-level confidence into topology learning effectively mitigates the propagation of ambiguous structural information, purifying feature aggregation.

\subsubsection{Adaptive Boundary}
Replacing class-specific adaptive boundaries with a unified threshold (denoted as 'Unif.Bo.') triggers the most severe performance plunge. This stark contrast highlights that different fault categories exhibit heterogeneous intra-class dispersion; the proposed Gaussian-based adaptive boundary is therefore the most critical component for reliable open-set rejection. 

\subsection{Parameter Sensitivity Analysis}

To investigate the robustness of the proposed framework, we conduct a parameter sensitivity analysis on the hyper-parameter crucial to the dual-criteria inference mechanism.  

The hyper-parameter $\tau_r$ dictates the trade-off between known fault identification and unknown fault rejection. As illustrated in Fig. \ref{fig:relative criterion}, we evaluate its impact on the H-score across three transfer tasks (P2, P4, P6) of the PU dataset. A relatively small $\tau_r$ (e.g., $0.1, 0.2$) imposes an overly strict rejection boundary, erroneously misclassifying ambiguous known samples as unknown ones. Conversely, when $\tau_r$ exceeds $0.4$, the criterion becomes excessively loose, allowing unknown faults with deceptive features to infiltrate the known classes, which triggers a sharp decline in overall performance. The optimal balance is consistently achieved at $\tau_r = 0.3$, thereby validating it as the default configuration.

\section{CONCLUSIONS} 

We propose a new PGU-OD framework to address open-set fault diagnosis under domain shifts via an end-to-end probabilistic closed-loop mechanism. PISA-Net leverages physically constrained wavelet convolutions to extract condition-robust features, mitigating perceptual uncertainty. The uncertainty-aware adaptive graph, governed by a heteroscedastic kernel, suppresses structural propagation of ambiguous information. The Gaussian-based boundary module optimizes intra-class compactness and enables dual-criteria inference for unknown fault rejection. Experiments confirm PGU-OD's robustness across domains. Future work includes continuous degradation modeling, few-shot open-set adaptation, and uncertainty calibration for industrial deployment.






\bibliographystyle{IEEEtran}
\bibliography{references}

@article{liu2025adaptive,
  title={Adaptive reconstruct feature difference network for open set domain generalization fault diagnosis},
  author={Liu, Mengyu and Cheng, Zhe and Yang, Yu and Hu, Niaoqing and Shen, Guoji and Yang, Yi},
  journal={Engineering Applications of Artificial Intelligence},
  volume={142},
  pages={109895},
  year={2025},
  publisher={Elsevier}
}

@article{chen2023bayesian,
  title={Bayesian hierarchical graph neural networks with uncertainty feedback for trustworthy fault diagnosis of industrial processes},
  author={Chen, Dongyue and Xie, Zongxia and Liu, Ruonan and Yu, Wenlong and Hu, Qinghua and Li, Xianling and Ding, Steven X},
  journal={IEEE Transactions on Neural Networks and Learning Systems},
  volume={35},
  number={12},
  pages={18635--18648},
  year={2023},
  publisher={IEEE}
}

@inproceedings{rehman2023open,
  title={Open set recognition methods for fault diagnosis: A review},
  author={Rehman, Attiq Ur and Jiao, Weidong and Sun, Jianfeng and Pan, Huilin and Yan, Tianyu},
  booktitle={2023 15th International Conference on Advanced Computational Intelligence (ICACI)},
  pages={1--8},
  year={2023},
  organization={IEEE}
}

@article{wang2025knowledge,
  title={Knowledge transfer and reinforcement based on biunbiased neural network: A novel solution for open-set fault transfer diagnosis},
  author={Wang, Lei and Zhang, Huaguang and Liu, Jinhai and Zuo, Fengyuan},
  journal={IEEE Transactions on Neural Networks and Learning Systems},
  year={2025},
  publisher={IEEE}
}

@article{chen2023transfer,
  title={Transfer learning-motivated intelligent fault diagnosis designs: A survey, insights, and perspectives},
  author={Chen, Hongtian and Luo, Hao and Huang, Biao and Jiang, Bin and Kaynak, Okyay},
  journal={IEEE Transactions on Neural Networks and Learning Systems},
  volume={35},
  number={3},
  pages={2969--2983},
  year={2023},
  publisher={IEEE}
}

@inproceedings{bendale2016towards,
  title={Towards open set deep networks},
  author={Bendale, Abhijit and Boult, Terrance E},
  booktitle={Proceedings of the IEEE conference on computer vision and pattern recognition},
  pages={1563--1572},
  year={2016}
}

@article{yu2021deep,
  title={Deep-learning-based open set fault diagnosis by extreme value theory},
  author={Yu, Xiaolei and Zhao, Zhibin and Zhang, Xingwu and Zhang, Qiyang and Liu, Yilong and Sun, Chuang and Chen, Xuefeng},
  journal={IEEE Transactions on Industrial Informatics},
  volume={18},
  number={1},
  pages={185--196},
  year={2021},
  publisher={IEEE}
}

@article{liao2025novel,
  title={A novel zero-shot learning method with feature generation for intelligent fault diagnosis},
  author={Liao, Wenjie and Wu, Like and Xu, Shihui and Fujimura, Shigeru},
  journal={IEEE Transactions on Industrial Informatics},
  volume={21},
  number={4},
  pages={3386--3395},
  year={2025},
  publisher={IEEE}
}

@article{jian2024open,
  title={Open-set domain generalization for fault diagnosis through data augmentation and a dual-level weighted mechanism},
  author={Jian, Chuanxia and Peng, Yonghe and Mo, Guopeng and Chen, Heen},
  journal={Advanced Engineering Informatics},
  volume={62},
  pages={102703},
  year={2024},
  publisher={Elsevier}
}

@article{Geng2025efecann, 
title={Domain Adaptation With Joint Distribution Alignment Adversarial Learning for Open-Set Bearing Intelligent Fault Diagnosis}, 
volume={25}, 
url={http://dx.doi.org/10.1109/jsen.2025.3576833}, 
DOI={10.1109/jsen.2025.3576833}, 
number={14}, 
journal={IEEE Sensors Journal}, 
publisher={Institute of Electrical and Electronics Engineers (IEEE)}, 
author={Geng, Yuanhao and Tang, Gang and Wang, Haoyang}, 
year={2025}, 
month={Jul}, 
pages={26507–26519} 
}

@ARTICLE{Geng2021OSFD,
  author={Geng, Chuanxing and Huang, Sheng-Jun and Chen, Songcan},
  journal={IEEE Transactions on Pattern Analysis and Machine Intelligence}, 
  title={Recent Advances in Open Set Recognition: A Survey}, 
  year={2021},
  volume={43},
  number={10},
  pages={3614-3631},
  doi={10.1109/TPAMI.2020.2981604}}

@article{Xu2026fosfd, 
title={Federated Open-Set Fault Diagnosis for Unknown Bearing Fault Detection}, 
volume={246}, 
url={http://dx.doi.org/10.1016/j.ymssp.2026.113917}, DOI={10.1016/j.ymssp.2026.113917}, 
journal={Mechanical Systems and Signal Processing}, 
publisher={Elsevier BV}, 
author={Xu, Danya and Jia, Mingwei and Chen, Tao and Liu, Yi and Chen, Dongyue and Chai, Tianyou and Yang, Tao}, 
year={2026}, 
month={Feb}, 
pages={113917}, 
language={en} 
}

@article{li2022explainable,
  title={Explainable graph wavelet denoising network for intelligent fault diagnosis},
  author={Li, Tianfu and Sun, Chuang and Li, Sinan and Wang, Zhiying and Chen, Xuefeng and Yan, Ruqiang},
  journal={IEEE Transactions on Neural Networks and Learning Systems},
  volume={35},
  number={6},
  pages={8535--8548},
  year={2022},
  publisher={IEEE}
}

@article{khosla2020supervised,
  title={Supervised contrastive learning},
  author={Khosla, Prannay and Teterwak, Piotr and Wang, Chen and Sarna, Aaron and Tian, Yonglong and Isola, Phillip and Maschinot, Aaron and Liu, Ce and Krishnan, Dilip},
  journal={Advances in neural information processing systems},
  volume={33},
  pages={18661--18673},
  year={2020}
}

@article{smith2015rolling,
  title={Rolling element bearing diagnostics using the Case Western Reserve University data: A benchmark study},
  author={Smith, Wade A and Randall, Robert B},
  journal={Mechanical systems and signal processing},
  volume={64},
  pages={100--131},
  year={2015},
  publisher={Elsevier}
}

@inproceedings{lessmeier2016condition,
  title={Condition monitoring of bearing damage in electromechanical drive systems by using motor current signals of electric motors: A benchmark data set for data-driven classification},
  author={Lessmeier, Christian and Kimotho, James Kuria and Zimmer, Detmar and Sextro, Walter},
  booktitle={PHM society European conference},
  volume={3},
  number={1},
  year={2016}
}

@article{lu2024novel,
  title={A novel multidomain contrastive-coding-based open-set domain generalization framework for machinery fault diagnosis},
  author={Lu, Biliang and Zhang, Yingjie and Sun, Qingshuai and Li, Ming and Li, Pude},
  journal={IEEE Transactions on Industrial Informatics},
  volume={20},
  number={4},
  pages={6369--6381},
  year={2024},
  publisher={IEEE}
}

@article{jia2025auxiliary,
  title={Auxiliary-feature-embedded causality-inspired dynamic penalty networks for open-set domain generalization diagnosis scenario},
  author={Jia, Ning and Huang, Weiguo and Ding, Chuancang and Huangfu, Yifan and Shi, Juanjuan and Zhu, Zhongkui},
  journal={Advanced Engineering Informatics},
  volume={65},
  pages={103220},
  year={2025},
  publisher={Elsevier}
}

@inproceedings{saito2018open,
  title={Open set domain adaptation by backpropagation},
  author={Saito, Kuniaki and Yamamoto, Shohei and Ushiku, Yoshitaka and Harada, Tatsuya},
  booktitle={Proceedings of the European conference on computer vision (ECCV)},
  pages={153--168},
  year={2018}
}

@article{zhao2022dual,
  title={Dual adversarial network for cross-domain open set fault diagnosis},
  author={Zhao, Chao and Shen, Weiming},
  journal={Reliability Engineering \& System Safety},
  volume={221},
  pages={108358},
  year={2022},
  publisher={Elsevier}
}

@article{zhao2022adaptive,
  title={Adaptive open set domain generalization network: Learning to diagnose unknown faults under unknown working conditions},
  author={Zhao, Chao and Shen, Weiming},
  journal={Reliability Engineering \& System Safety},
  volume={226},
  pages={108672},
  year={2022},
  publisher={Elsevier}
}

@article{1,
  title={Intelligent fault diagnosis for large-scale rotating machines using binarized deep neural networks and random forests},
  author={Li, Huifang and Hu, Guangzheng and Li, Jianqiang and Zhou, Mengchu},
  journal={IEEE Transactions on Automation Science and Engineering},
  volume={19},
  number={2},
  pages={1109--1119},
  year={2021},
  publisher={IEEE}
}

@article{2,
  title={A discrimination-guided active learning method based on marginal representations for industrial compound fault diagnosis},
  author={Liu, Zeyi and Zhang, Jingfei and He, Xiao},
  journal={IEEE Transactions on Automation Science and Engineering},
  volume={21},
  number={4},
  pages={6411--6422},
  year={2023},
  publisher={IEEE}
}

@article{3,
  title={A transactional-behavior-based hierarchical gated network for credit card fraud detection},
  author={Xie, Yu and Zhou, MengChu and Liu, Guanjun and Wei, Lifei and Zhu, Honghao and De Meo, Pasquale},
  journal={IEEE/CAA Journal of Automatica Sinica},
  year={2025},
  publisher={IEEE}
}

@article{4,
  title={A Multi-objective Optimization Approach for Feature Selection in Gentelligent Systems},
  author={Ghahramani, Mohammadhossein and Qiao, Yan and Wu, NaiQi and Zhou, Mengchu},
  journal={IEEE Internet of Things Journal},
  year={2025},
  publisher={IEEE}
}

@article{wang2026advanced,
  title={Advanced high-order graph convolutional networks with assorted time-frequency transforms},
  author={Wang, Ling and Yuan, Ye and Luo, Xin},
  journal={IEEE/CAA Journal of Automatica Sinica},
  pages={394--408},
  year={2026},
  publisher={IEEE}
}

@inproceedings{G3,
  title={Sgd-dyg: Self-reliant global dependency apprehending on dynamic graphs},
  author={Han, Minglian and Wang, Ling and Yuan, Ye and Luo, Xin},
  booktitle={Proceedings of the 31st ACM SIGKDD Conference on Knowledge Discovery and Data Mining V. 2},
  pages={802--813},
  year={2025}
}

@article{G4,
  title={GT-A 2 T: Graph tensor alliance attention network},
  author={Wang, Ling and Liu, Kechen and Yuan, Ye},
  journal={IEEE/CAA Journal of Automatica Sinica},
  volume={12},
  number={10},
  pages={2165--2167},
  year={2024},
  publisher={IEEE}
}

@article{G5,
  title={A node-collaboration-informed graph convolutional network for highly accurate representation to undirected weighted graph},
  author={Yuan, Ye and Wang, Ying and Luo, Xin},
  journal={IEEE Transactions on Neural Networks and Learning Systems},
  volume={36},
  number={6},
  pages={11507--11519},
  year={2024},
  publisher={IEEE}
}

@article{G6,
  title={Adaptive pid-incorporated nonnegative latent factor analysis},
  author={Li, Jinli and Yuan, Ye and He, Tiantian and Luo, Xin},
  journal={IEEE Transactions on Systems, Man, and Cybernetics: Systems},
  year={2026},
  publisher={IEEE}
}

@article{G7,
  title={Learning error refinement in stochastic gradient descent-based latent factor analysis via diversified pid controllers},
  author={Li, Jinli and Yuan, Ye and Luo, Xin},
  journal={IEEE Transactions on Emerging Topics in Computational Intelligence},
  year={2025},
  publisher={IEEE}
}

@article{G8,
  title={A Kalman-filter-incorporated latent factor analysis model for temporally dynamic sparse data},
  author={Yuan, Ye and Luo, Xin and Shang, Mingsheng and Wang, Zidong},
  journal={IEEE Transactions on Cybernetics},
  volume={53},
  number={9},
  pages={5788--5801},
  year={2022},
  publisher={IEEE}
}

@inproceedings{G9,
  title={A generalized and fast-converging non-negative latent factor model for predicting user preferences in recommender systems},
  author={Yuan, Ye and Luo, Xin and Shang, Mingsheng and Wu, Di},
  booktitle={Proceedings of The Web Conference 2020},
  pages={498--507},
  year={2020}
}

@article{G10,
  title={Tensor low-rank orthogonal compression for convolutional neural networks},
  author={He, Yaping and Luo, Xin},
  journal={IEEE/CAA Journal of Automatica Sinica},
  volume={13},
  number={1},
  pages={227--229},
  year={2026},
  publisher={IEEE}
}

@article{bi2025discovering,
  title={Discovering spatiotemporal--individual coupled features from nonstandard tensors—a novel dynamic graph mixer approach},
  author={Bi, Fanghui and He, Tiantian and Ong, Yew-Soon and Luo, Xin},
  journal={IEEE Transactions on Neural Networks and Learning Systems},
  year={2025},
  publisher={IEEE}
}

@article{wang2026graph,
  title={Graph tensor convolutional network},
  author={Wang, Ling and Yuan, Ye and Luo, Xin},
  journal={IEEE Transactions on Systems, Man, and Cybernetics: Systems},
  year={2026},
  publisher={IEEE}
}

@article{he2026modularized,
  title={Modularized graph convolutional network},
  author={He, Tiantian and Duan, Zhixuan and Luo, Xin},
  journal={IEEE/CAA Journal of Automatica Sinica},
  volume={13},
  number={3},
  pages={737--739},
  year={2026},
  publisher={IEEE}
}

@article{bi2024graph,
  title={Graph linear convolution pooling for learning in incomplete high-dimensional data},
  author={Bi, Fanghui and He, Tiantian and Ong, Yew-Soon and Luo, Xin},
  journal={IEEE Transactions on Knowledge and Data Engineering},
  volume={37},
  number={4},
  pages={1838--1852},
  year={2024},
  publisher={IEEE}
}

@article{liu2023symmetry,
  title={Symmetry and graph bi-regularized non-negative matrix factorization for precise community detection},
  author={Liu, Zhigang and Luo, Xin and Zhou, Mengchu},
  journal={IEEE Transactions on Automation Science and Engineering},
  volume={21},
  number={2},
  pages={1406--1420},
  year={2023},
  publisher={IEEE}
}

@article{yuan2025proportional,
  title={A proportional integral controller-enhanced non-negative latent factor analysis model},
  author={Yuan, Ye and Lu, Siyang and Luo, Xin},
  journal={IEEE/CAA Journal of Automatica Sinica},
  volume={12},
  number={6},
  pages={1246--1259},
  year={2025},
  publisher={IEEE}
}

@article{yuan2024fuzzy,
  title={A fuzzy PID-incorporated stochastic gradient descent algorithm for fast and accurate latent factor analysis},
  author={Yuan, Ye and Li, Jinli and Luo, Xin},
  journal={IEEE Transactions on Fuzzy Systems},
  volume={32},
  number={7},
  pages={4049--4061},
  year={2024},
  publisher={IEEE}
}

@article{yuan2023adaptive,
  title={An adaptive divergence-based non-negative latent factor model},
  author={Yuan, Ye and Wang, Renfang and Yuan, Guangxiao and Xin, Luo},
  journal={IEEE Transactions on Systems, Man, and Cybernetics: Systems},
  volume={53},
  number={10},
  pages={6475--6487},
  year={2023},
  publisher={IEEE}
}

@inproceedings{yuan2020temporal,
  title={Temporal Web Service QoS Prediction via Kalman Filter-Incorporated Latent Factor Analysis.},
  author={Yuan, Ye and Shang, Mingsheng and Luo, Xin},
  booktitle={ECAI},
  pages={561--568},
  year={2020}
}

@article{qin2026robust,
  title={A robust approach to electricity theft detection via tensor representation-driven contrastive distillation},
  author={Qin, Wen and Ding, Yuting and Luo, Xin},
  journal={IEEE Transactions on Industrial Informatics},
  year={2026},
  publisher={IEEE}
}

@article{hu2025comprehensive,
  title={A Comprehensive Review of Parallel Optimization Algorithms for High-Dimensional and Incomplete Matrix Factorization},
  author={Hu, Qicong and Wu, Hao and Luo, Xin},
  journal={IEEE/CAA Journal of Automatica Sinica},
  volume={12},
  number={12},
  pages={2399--2426},
  year={2025},
  publisher={IEEE}
}

@article{chen2024latent,
  title={Latent-factorization-of-tensors-incorporated battery cycle life prediction},
  author={Chen, Minzhi and Tao, Li and Lou, Jungang and Luo, Xin},
  journal={IEEE/CAA Journal of Automatica Sinica},
  volume={12},
  number={3},
  pages={633--635},
  year={2024},
  publisher={IEEE}
}

@article{wu2023robust,
  title={Robust low-rank latent feature analysis for spatiotemporal signal recovery},
  author={Wu, Di and Li, Zechao and Yu, Zhikai and He, Yi and Luo, Xin},
  journal={IEEE Transactions on Neural Networks and Learning Systems},
  volume={36},
  number={2},
  pages={2829--2842},
  year={2023},
  publisher={IEEE}
}

@article{chen2024generalized,
  title={A generalized nesterov's accelerated gradient-incorporated non-negative latent-factorization-of-tensors model for efficient representation to dynamic QoS data},
  author={Chen, Minzhi and Wang, Renfang and Qiao, Yan and Luo, Xin},
  journal={IEEE Transactions on Emerging Topics in Computational Intelligence},
  volume={8},
  number={3},
  pages={2386--2400},
  year={2024},
  publisher={IEEE}
}

@article{yuan2026novel,
  title={A novel approach to temporal qos estimation via extended kalman filter-incorporated latent feature analysis},
  author={Yuan, Ye and Wang, Song and Zhou, Hongxun and Wang, Ling and Luo, Xin},
  journal={IEEE Transactions on Services Computing},
  year={2026},
  publisher={IEEE}
}

@article{shang2018randomized,
  title={Randomized latent factor model for high-dimensional and sparse matrices from industrial applications},
  author={Shang, Mingsheng and Luo, Xin and Liu, Zhigang and Chen, Jia and Yuan, Ye and Zhou, MengChu},
  journal={IEEE/CAA Journal of Automatica Sinica},
  volume={6},
  number={1},
  pages={131--141},
  year={2018},
  publisher={IEEE}
}

@article{lin2025ncsac,
  title={Ncsac: Effective neural community search via attribute-augmented conductance},
  author={Lin, Longlong and Li, Quanao and Qiao, Miao and Wang, Zeli and Zhao, Jin and Li, Rong-Hua and Luo, Xin and Jia, Tao},
  journal={IEEE Transactions on Knowledge and Data Engineering},
  volume={38},
  number={2},
  pages={1221--1235},
  year={2025},
  publisher={IEEE}
}

@article{xu2025attention,
  title={Attention-mechanism-based neural latent-factorization-of-tensors model},
  author={Xu, Xiuqin and Lin, Mingwei and Xu, Zeshui and Luo, Xin},
  journal={ACM Transactions on Knowledge Discovery from Data},
  volume={19},
  number={4},
  pages={1--27},
  year={2025},
  publisher={ACM New York, NY}
}

@article{wu2023mmlf,
  title={Mmlf: Multi-metric latent feature analysis for high-dimensional and incomplete data},
  author={Wu, Di and Zhang, Peng and He, Yi and Luo, Xin},
  journal={IEEE transactions on services computing},
  volume={17},
  number={2},
  pages={575--588},
  year={2023},
  publisher={IEEE}
}

@article{li2025neural,
  title={Neural nonnegative latent factorization of tensors model with acceleration and unconstraint},
  author={Li, Wenqiang and Lin, Mingwei and Xu, Xiuqin and Lin, Ling and Xu, Zeshui and Luo, Xin},
  journal={IEEE Transactions on Systems, Man, and Cybernetics: Systems},
  year={2025},
  publisher={IEEE}
}

@article{qin2023asynchronous,
  title={Asynchronous parallel fuzzy stochastic gradient descent for high-dimensional incomplete data representation},
  author={Qin, Wen and Luo, Xin},
  journal={IEEE Transactions on Fuzzy Systems},
  volume={32},
  number={2},
  pages={445--459},
  year={2023},
  publisher={IEEE}
}

@article{wang2024distributed,
  title={A distributed adaptive second-order latent factor analysis model},
  author={Wang, Jialiang and Li, Weiling and Luo, Xin},
  journal={IEEE/CAA Journal of Automatica Sinica},
  volume={11},
  number={11},
  pages={2343--2345},
  year={2024},
  publisher={IEEE}
}

@article{wu2026multimetric,
  title={Multimetric autoencoder for representing high-dimensional and incomplete data},
  author={Wu, Di and Liang, Cheng and He, Yi and Qiao, Yan and Luo, Xin},
  journal={IEEE Transactions on Systems, Man, and Cybernetics: Systems},
  volume={56},
  number={3},
  pages={1533--1546},
  year={2026},
  publisher={IEEE}
}

@article{lyu2025dynamic,
  title={Dynamic stochastic reorientation particle swarm optimization for adaptive latent factor analysis in high-dimensional sparse matrices},
  author={Lyu, Chao and Ma, Ziwen and Luo, Xin and Shi, Yuhui},
  journal={IEEE Transactions on Knowledge and Data Engineering},
  year={2025},
  publisher={IEEE}
}

@ARTICLE{lyu2026gnetic,
  author={Lyu, Chao and Cheng, Jingna and Luo, Xin and Shi, Yuhui},
  journal={IEEE Transactions on Neural Networks and Learning Systems}, 
  title={Genetic Algorithm-Based Two-Step Optimization for Precise Latent Factor Analysis}, 
  year={2026},
  volume={37},
  publisher={IEEE}
  }

@ARTICLE{qin2024parallel,
  author={Qin, Wen and Luo, Xin and Li, Shuai and Zhou, MengChu},
  journal={IEEE Transactions on Automation Science and Engineering}, 
  title={Parallel Adaptive Stochastic Gradient Descent Algorithms for Latent Factor Analysis of High-Dimensional and Incomplete Industrial Data}, 
  year={2024},
  volume={21},
  publisher={IEEE}
  }

@article{jiang2024iterative,
  title={Iterative role negotiation via the bilevel GRA++ with decision tolerance},
  author={Jiang, Qian and Liu, Dongning and Zhu, Haibin and Wu, Shijue and Wu, Naiqi and Luo, Xin and Qiao, Yan},
  journal={IEEE Transactions on Computational Social Systems},
  volume={11},
  number={6},
  pages={7484--7499},
  year={2024},
  publisher={IEEE}
}

@ARTICLE{xu2026sampling,
  author={Xu, Xiuqin and Lin, Mingwei and Xu, Zeshui and Luo, Xin},
  journal={IEEE Transactions on Network and Service Management}, 
  title={A Sampling-Neighborhood-Regularized Latent Factorization of Tensor for Dynamic QoS Estimation}, 
  year={2026},
  volume={23},
  pubisher={IEEE}
  }

@article{liao2025local,
  title={Local search-based anytime algorithms for continuous distributed constraint optimization problems},
  author={Liao, Xin and Hoang, Khoi and Luo, Xin},
  journal={IEEE/CAA Journal of Automatica Sinica},
  volume={12},
  number={1},
  pages={288--290},
  year={2025},
  publisher={IEEE}
}

@ARTICLE{bi2024fast,
  author={Bi, Fanghui and He, Tiantian and Luo, Xin},
  journal={IEEE Transactions on Services Computing}, 
  title={A Fast Nonnegative Autoencoder-Based Approach to Latent Feature Analysis on High-Dimensional and Incomplete Data}, 
  year={2024},
  volume={17},
  number={3},
  pages={733-746},
  publisher={IEEE}
}

@article{wu2026non,
  title={Non-Gradient Hash Factor Learning for High-Dimensional and Incomplete Data Representation Learning},
  author={Wu, Di and Li, Shihui and He, Yi and Luo, Xin and Gao, Xinbo},
  journal={IEEE Transactions on Pattern Analysis and Machine Intelligence},
  year={2026},
  publisher={IEEE}
}

@article{tang2025auto,
  title={Auto-encoding neural tucker factorization},
  author={Tang, Peng and Luo, Xin and Woodcock, Jim},
  journal={IEEE Transactions on Knowledge and Data Engineering},
  year={2025},
  publisher={IEEE}
}

@article{wang2025convolution,
  title={A convolution bias-incorporated nonnegative latent factorization of tensors model for accurate representation learning to dynamic directed graphs},
  author={Wang, Qu and Wu, Hao and Luo, Xin},
  journal={IEEE Transactions on Systems, Man, and Cybernetics: Systems},
  year={2025},
  publisher={IEEE}
}

@article{wu2024fine,
  title={A fine-grained regularization scheme for non-negative latent factorization of high-dimensional and incomplete tensors},
  author={Wu, Hao and Qiao, Yan and Luo, Xin},
  journal={IEEE Transactions on Services Computing},
  volume={17},
  number={6},
  pages={3006--3021},
  year={2024},
  publisher={IEEE}
}

@ARTICLE{tang2025nerual,
  author={Tang, Peng and Luo, Xin},
  journal={IEEE/CAA Journal of Automatica Sinica}, 
  title={Neural Tucker Factorization}, 
  year={2025},
  volume={12},
  number={2},
  publisher={IEEE}
  }

@article{wei2022robust,
  title={A robust coevolutionary neural-based optimization algorithm for constrained nonconvex optimization},
  author={Wei, Lin and Jin, Long and Luo, Xin},
  journal={IEEE Transactions on Neural Networks and Learning Systems},
  volume={35},
  number={6},
  pages={7778--7791},
  year={2022},
  publisher={IEEE}
}

@ARTICLE{yang2025link,
  author={Yang, Yue and Hu, Lun and Li, Guodong and Li, Dongxu and Hu, Pengwei and Luo, Xin},
  journal={IEEE Transactions on Systems, Man, and Cybernetics: Systems}, 
  title={Link-Based Attributed Graph Clustering via Approximate Generative Bayesian Learning}, 
  year={2025},
  volume={55},
  number={8},
  pages={5730-5743},  
  publisher={IEEE}
}

@ARTICLE{deng2026fuzzy,
  author={Deng, Xun and Hu, Pengwei and Herget, Thomas and Tan, Feng and Zhu, Xiaobo and Zhang, Jun and Huang, Yu-an and Hu, Lun and You, Zhuhong and Luo, Xin},
  journal={IEEE Transactions on Fuzzy Systems}, 
  title={Fuzzy Mixture-of-Experts Aggregation for Organoid Identification With Multiscale State Space Features}, 
  year={2026},
  volume={34},
  number={1},
  pages={324-335},
  publisher={IEEE}
}

@ARTICLE{qin2024ada,
  author={Qin, Wen and Luo, Xin and Zhou, MengChu},
  journal={IEEE Transactions on Big Data},
  title={Adaptively-Accelerated Parallel Stochastic Gradient Descent for High-Dimensional and Incomplete Data Representation Learning},
  year={2024},
  volume={10},
  number={1},
  pages={92-107},
  publisher={IEEE}
}

@article{jiang2025dual,
  title={Dual graph driven-consistent representation learning method for semi-supervised fault diagnosis of rotating machinery},
  author={Jiang, Zhichao and Liu, Dongdong and Wang, Huaqing and Cui, Lingli},
  journal={Advanced Engineering Informatics},
  volume={65},
  pages={103274},
  year={2025},
  publisher={Elsevier},
  doi={10.1016/j.aei.2025.103274}
}

@article{sun2025multi,
  title={Multi-sensor temporal-spatial graph network fusion empirical mode decomposition convolution for machine fault diagnosis},
  author={Sun, Kuangchi and Yin, Aijun},
  journal={Information Fusion},
  volume={114},
  pages={102708},
  year={2025},
  publisher={Elsevier},
  doi={10.1016/j.inffus.2024.102708}
}

@inproceedings{liu2024pretraining,
  title={Pre-training graph neural network for fault diagnosis and safety assessment},
  author={Liu, Chang and He, Xiao and Dong, Hairong},
  booktitle={2024 IEEE International Conference on Industrial Technology (ICIT)},
  pages={1--6},
  year={2024},
  organization={IEEE},
  doi={10.1109/ICIT58233.2024.10540872}
}

@article{fu2025reason,
  title={Reason and discovery: A new paradigm for open set recognition},
  author={Fu, Yimin and Liu, Zhunga and Lyu, Jialin},
  journal={IEEE Transactions on Pattern Analysis and Machine Intelligence},
  volume={47},
  number={7},
  pages={5586--5599},
  year={2025},
  publisher={IEEE},
  doi={10.1109/TPAMI.2025.3552760}
}

@article{fu2023magva,
  title={MAGVA: An open-set fault diagnosis model based on multi-hop attentive graph variational autoencoder for autonomous vehicles},
  author={Fu, Rao and Bi, Yuanguo and Han, Guangjie and Zhang, Xiaoling and Liu, Li and Zhao, Liang and Hu, Bing},
  journal={IEEE Transactions on Intelligent Transportation Systems},
  volume={24},
  number={12},
  pages={14873--14889},
  year={2023},
  publisher={IEEE},
  doi={10.1109/TITS.2023.3300911}
}

@article{xiao2025uncertainty,
  title={Uncertainty-aware deep variational attention network: A trustworthy mechanical fault diagnostic model assisted by out-of-distribution detection},
  author={Xiao, Yiming and Shao, Haidong and Zhang, Haomiao and Wei, Rongming and Liu, Bin},
  journal={Engineering Applications of Artificial Intelligence},
  volume={157},
  pages={111386},
  year={2025},
  publisher={Elsevier},
  doi={10.1016/j.engappai.2025.111386}
}

@article{he2023physics,
  title={Physics-informed interpretable wavelet weight initialization and balanced dynamic adaptive threshold for intelligent fault diagnosis of rolling bearings},
  author={He, Chao and Shi, Hongmei and Si, Jin and Li, Jianbo},
  journal={Journal of Manufacturing Systems},
  volume={70},
  pages={579--592},
  year={2023},
  publisher={Elsevier},
  doi={10.1016/j.jmsy.2023.08.014}
}

@article{li2025deep,
  title={Deep complex wavelet denoising network for interpretable fault diagnosis of industrial robots with noise interference and imbalanced data},
  author={Li, Rourou and Xia, Tangbin and Jiang, Yimin and Wu, Jianhua and Fang, Xiaolei and Gebraeel, Nagi and Xi, Lifeng},
  journal={IEEE Transactions on Instrumentation and Measurement},
  volume={74},
  pages={1--11},
  year={2025},
  publisher={IEEE},
  doi={10.1109/TIM.2025.3540131}
}

@article{wan2024self,
  title={Self-supervised simple Siamese framework for fault diagnosis of rotating machinery with unlabeled samples},
  author={Wan, Wenqing and Chen, Jinglong and Zhou, Zitong and Shi, Zhen},
  journal={IEEE Transactions on Neural Networks and Learning Systems},
  volume={35},
  number={5},
  pages={6380--6392},
  year={2024},
  publisher={IEEE},
  doi={10.1109/TNNLS.2022.3209332}
}

@article{jia2024causal,
  title={Causal disentanglement domain generalization for time-series signal fault diagnosis},
  author={Jia, Linshan and Chow, Tommy W.S. and Yuan, Yixuan},
  journal={Neural Networks},
  volume={172},
  pages={106099},
  year={2024},
  publisher={Elsevier},
  doi={10.1016/j.neunet.2024.106099}
}

@article{he2025individual,
  title={An individual generalization framework based on independent samples towards a more reasonable fault diagnosis benchmark},
  author={He, Yiming and Shen, Weiming},
  journal={Computers in Industry},
  volume={173},
  pages={104359},
  year={2025},
  publisher={Elsevier},
  doi={10.1016/j.compind.2025.104359}
}

@article{qian2025adaptive,
  title={Adaptive intermediate class-wise distribution alignment: A universal domain adaptation and generalization method for machine fault diagnosis},
  author={Qian, Quan and Luo, Jun and Qin, Yi},
  journal={IEEE Transactions on Neural Networks and Learning Systems},
  volume={36},
  number={3},
  pages={4296--4310},
  year={2025},
  publisher={IEEE},
  doi={10.1109/TNNLS.2024.3376449}
}

@article{shao2025lsfconvformer,
  title={LSFConvformer: A lightweight method for mechanical fault diagnosis under small samples and variable speeds with time-frequency fusion},
  author={Shao, Haidong and Lai, Yanzuo and Liu, Haoran and Wang, Jie and Liu, Bin},
  journal={Mechanical Systems and Signal Processing},
  volume={236},
  pages={113016},
  year={2025},
  publisher={Elsevier},
  doi={10.1016/j.ymssp.2025.113016}
}

@article{tang2025complex,
  title={A complex attention transformer for bearing fault diagnosis based on motor current signals},
  author={Tang, Jiayin and Zhou, Qiuyang and Yin, Kexin},
  journal={IEEE Transactions on Instrumentation and Measurement},
  volume={74},
  pages={1--11},
  year={2025},
  publisher={IEEE},
  doi={10.1109/TIM.2025.3571168}
}

@article{magadan2024explainable,
  title={Explainable and interpretable bearing fault classification and diagnosis under limited data},
  author={Magad{\'a}n, L. and Ruiz-C{\'a}rcel, C. and Granda, J.C. and Su{\'a}rez, F.J. and Starr, A.},
  journal={Advanced Engineering Informatics},
  volume={62},
  pages={102909},
  year={2024},
  publisher={Elsevier},
  doi={10.1016/j.aei.2024.102909}
}

@inproceedings{chaouech2025advanced,
  title={Advanced feature extraction techniques for bearing fault diagnosis using higher-order statistics and machine learning},
  author={Chaouech, Lotfi and Ali, Jaouher Ben},
  booktitle={2025 International Conference on Control, Automation and Diagnosis (ICCAD)},
  pages={1--6},
  year={2025},
  organization={IEEE},
  doi={10.1109/ICCAD64771.2025.11099218}
}

@article{chang2025causal,
  title={Causal disentanglement-based hidden Markov model for cross-domain bearing fault diagnosis},
  author={Chang, Rihao and Ma, Yongtao and Nie, Weizhi and Nie, Jie and Zhu, Yiqun and Liu, An-An},
  journal={IEEE Transactions on Neural Networks and Learning Systems},
  volume={36},
  number={8},
  pages={13968--13982},
  year={2025},
  publisher={IEEE},
  doi={10.1109/TNNLS.2024.3513329}
}

@article{yan2025compact,
  title={Compact-sparse prototype calibration network for few-shot continual fault diagnosis of rotating machinery},
  author={Yan, Shen and Shao, Haidong and Wang, Xinyi and Zhang, Haomiao and Xiao, Yiming and Liu, Bin},
  journal={Engineering Applications of Artificial Intelligence},
  volume={155},
  pages={111099},
  year={2025},
  publisher={Elsevier},
  doi={10.1016/j.engappai.2025.111099}
}

@article{shao2025safs,
  title={SAFS-Net: A novel health indicator extraction and fault early warning method for machinery},
  author={Shao, Minghui and Shao, Haidong and Feng, Minjie and Yan, Shen and Liu, Bin},
  journal={Advanced Engineering Informatics},
  volume={68},
  pages={103614},
  year={2025},
  publisher={Elsevier},
  doi={10.1016/j.aei.2025.103614}
}

@article{shao2021novel,
  title={A novel approach of multisensory fusion to collaborative fault diagnosis in maintenance},
  author={Shao, Haidong and Lin, Jing and Zhang, Liangwei and Galar, Diego and Kumar, Uday},
  journal={Information Fusion},
  volume={74},
  pages={65--76},
  year={2021},
  publisher={Elsevier},
  doi={10.1016/j.inffus.2021.03.008}
}

@article{shao2021intelligent,
  title={Intelligent fault diagnosis of rotor-bearing system under varying working conditions with modified transfer convolutional neural network and thermal images},
  author={Shao, Haidong and Xia, Min and Han, Guangjie and Zhang, Yu and Wan, Jiafu},
  journal={IEEE Transactions on Industrial Informatics},
  volume={17},
  number={5},
  pages={3488--3496},
  year={2021},
  publisher={IEEE},
  doi={10.1109/TII.2020.3005965}
}

@article{zhao2020deep,
  title={Deep learning algorithms for rotating machinery intelligent diagnosis: An open source benchmark study},
  author={Zhao, Zhibin and Li, Tianfu and Wu, Jingyao and Sun, Chuang and Wang, Shibin and Yan, Ruqiang and Chen, Xuefeng},
  journal={ISA Transactions},
  volume={107},
  pages={224--255},
  year={2020},
  publisher={Elsevier},
  doi={10.1016/j.isatra.2020.08.010}
}

@article{li2025openset,
  title={Open-set fault diagnosis in multimode processes via fine-grained deep feature representation},
  author={Li, Guangqiang and Atoui, M. Amine and Li, Xiangshun},
  journal={Process Safety and Environmental Protection},
  volume={204},
  pages={108106},
  year={2025},
  publisher={Elsevier},
  doi={10.1016/j.psep.2025.108106}
}

\end{document}